\documentclass[conference]{IEEEtran}

\usepackage{times}
\usepackage[numbers]{natbib}
\usepackage{epsfig}
\usepackage{graphicx}
\usepackage{amsmath}
\usepackage{amssymb}
\usepackage{algorithm,algorithmic}
\usepackage{multirow}
\usepackage{subcaption}
\usepackage{xcolor,colortbl}

\usepackage[pagebackref=true,breaklinks=true,colorlinks,bookmarks=true]{hyperref}

\begin{document}

\title{\Large\bf ReAgent: Point Cloud Registration using Imitation and Reinforcement Learning}
\author{Dominik Bauer, Timothy Patten and Markus Vincze\\
TU Wien\\
Vienna, Austria\\
{\tt\small \{bauer,patten,vincze\}@acin.tuwien.ac.at}
}

\maketitle

\begin{abstract}
Point cloud registration is a common step in many 3D computer vision tasks such as object pose estimation, where a 3D model is aligned to an observation. Classical registration methods generalize well to novel domains but fail when given a noisy observation or a bad initialization. Learning-based methods, in contrast, are more robust but lack in generalization capacity. 
We propose to consider iterative point cloud registration as a reinforcement learning task and, to this end, present a novel registration agent (ReAgent). We employ imitation learning to initialize its discrete registration policy based on a steady expert policy. Integration with policy optimization, based on our proposed alignment reward, further improves the agent's registration performance. 
We compare our approach to classical and learning-based registration methods on both ModelNet40 (synthetic) and ScanObjectNN (real data) and show that our ReAgent achieves state-of-the-art accuracy. The lightweight architecture of the agent, moreover, enables reduced inference time as compared to related approaches. In addition, we apply our method to the object pose estimation task on real data (LINEMOD), outperforming state-of-the-art pose refinement approaches.
Code is available at \href{https://www.github.com/dornik/reagent}{github.com/dornik/reagent}.
\end{abstract}



\section{Introduction}
Depending on the application domain, point cloud registration methods need to fulfill a range of properties. For example, AR applications and robotics applications require real-time inference speed and robustness to unexpected observations. In such real-world deployment, generalization to categories that were not seen during training is required. Further, registration approaches also need to generalize to different tasks, such as object pose estimation or scan alignment. Finally, an interaction with or scrutiny by a human might be required. For this, the method's steps need to be interpretable. These properties are often competing and thus difficult to achieve using a single approach.

As diverse as the required properties are the approaches that are proposed to solve point cloud registration. Distinctive features of proposed methods are global \cite{yang2015goicp,zhou2016fgr} or local optimality \cite{besl1992icp}, as well as one-shot \cite{wang2019dcp} or iterative computation~\cite{aoki2019pointnetlk}. Global considerations allow for greater robustness to initial conditions than local methods, albeit at the cost of significantly increased computation time. While iterative methods may achieve higher accuracy than one-shot approaches through repeated registration, they may diverge over multiple steps.  Furthermore, learning-based approaches are proposed in related work \cite{wang2019dcp,aoki2019pointnetlk,wang2019prnet,choy2020dgr}, which are shown to be more robust to initialization and noise than classical approaches. However, these methods are not robust to domain change, e.g., when transferred to novel tasks.

\begin{figure}[t]
    \centering
    \includegraphics[width=\linewidth]{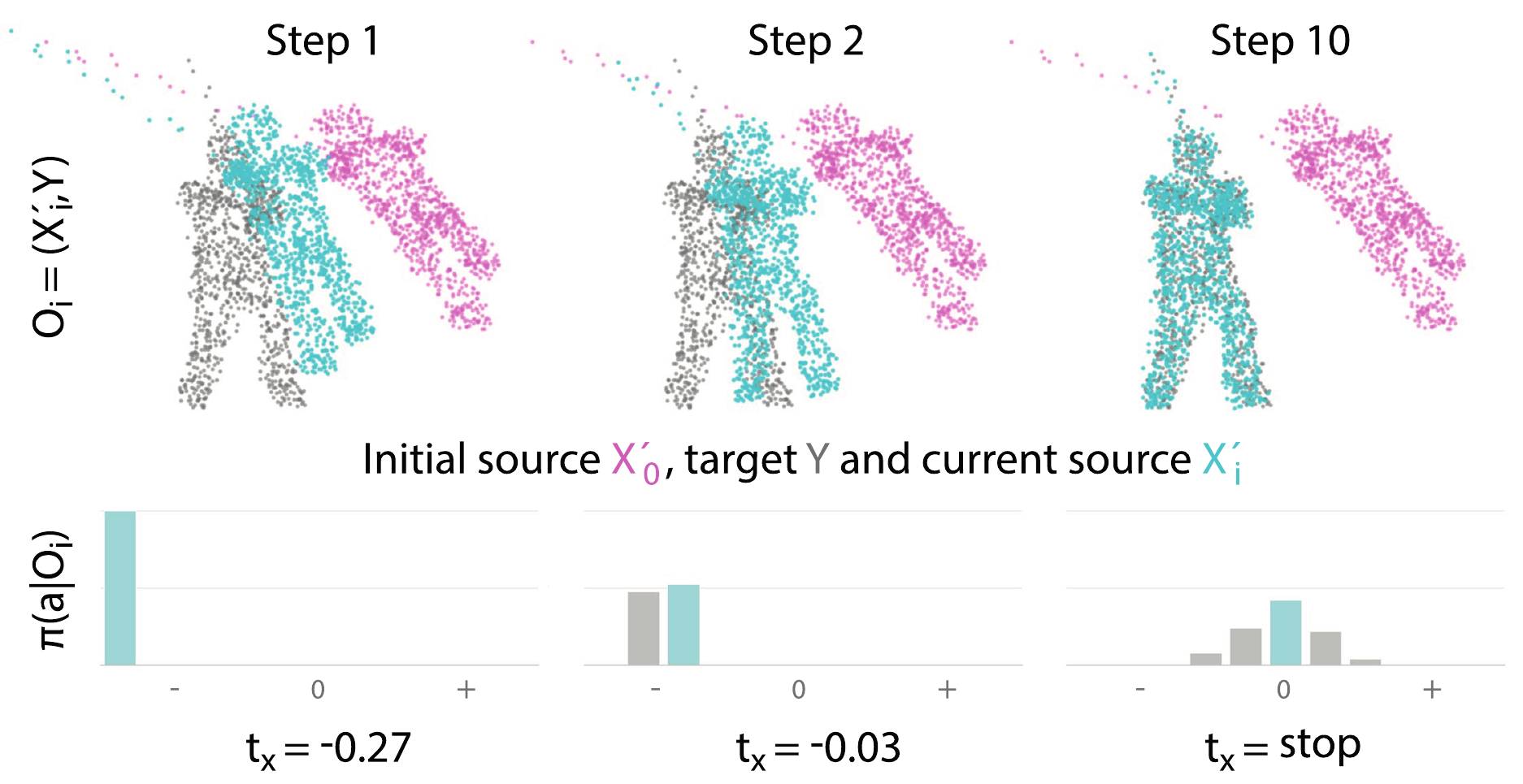}
    \caption{Iterative registration using ReAgent. The source point cloud (cyan) is aligned to the target point cloud (gray), starting from an initial source (magenta). ReAgent follows policy $\pi$ by taking action $a_i=\arg\max_a \pi(a|O_i)$ given the current observation $O_i$, improving registration step-by-step.}
    \label{fig:teaser}
\end{figure}

In an effort to bridge this gap between methods, we design a novel registration approach that unifies accuracy, robustness to noise and initialization with inference speed. While reinforcement learning methods for RGB-based object pose refinement are proposed \cite{shao2020pfrl,busam2020moveit}, to the best of our knowledge, we are the first to consider 3D point cloud registration as a reinforcement learning problem. Our approach is based on a combination of Imitation Learning (IL) and Reinforcement Learning (RL); imitating an expert to learn an accurate initial policy, reinforcing a symmetry-invariant reward to further improve the policy. The proposed registration agent (\textit{ReAgent}) treats registration as an iterative classification of the observed point cloud pair into discrete steps, as shown in Figure \ref{fig:teaser}. We
\begin{itemize}
    \item propose a combined imitation and reinforcement learning approach to point cloud registration,
    \item improve accuracy compared to state-of-the-art registration methods on synthetic and real data,
    \item improve accuracy compared to state-of-the-art pose refinement methods on LINEMOD and
    \item reduce inference time on both tasks using our lightweight agent as compared to related approaches.
\end{itemize}

We discuss related point cloud registration and reinforcement learning methods in Section \ref{sec:related}. Section \ref{sec:method} presents the proposed point cloud registration agent. Section \ref{sec:results} provides experiments on synthetic and real data and discusses the impact of the presented contributions. Section \ref{sec:conclusion} concludes the paper and indicates potential future work.


\section{Related Work}\label{sec:related}
The presented approach is influenced by previous work in point cloud registration and work that applies reinforcement learning to the related task of object pose estimation.

\textbf{Classical Point Cloud Registration:} The most influential work in point cloud registration is Iterative Closest Point (ICP) \cite{besl1992icp}. First, the (closest) points in the source and target points clouds are matched. Then, a transformation that minimizes the error between the matched points is found in closed form. These steps are repeated until convergence. Many variants are proposed, e.g., considering surface normals \cite{rusinkiewicz2001icp}, color \cite{park2017icp} or using non-linear optimization \cite{fitzgibbon2003icp}. For a detailed overview of ICP-based methods, see \cite{rusinkiewicz2001icp,tam2012icp}.

ICP may only find local optima. A branch-and-bound variant \cite{yang2015goicp} trades global optimality for increased runtime. Other global approaches use local features \cite{rusu2009fpfh} with RANSAC or directly optimize a global objective \cite{zhou2016fgr}. TEASER \cite{yang2020teaser} achieves high robustness to large amounts of outliers through a truncated least squares cost and allows to certify global optimality of the estimated registration.

\textbf{Learning-based Point Cloud Registration:} Recent approaches based on neural networks (NN) use the idea of ICP and its global variants. Local features are extracted to determine a matching between the input clouds. Using this matching, the transformation is found either in closed form using differentiable Weighted SVD \cite{wang2019dcp,wang2019prnet,yew2020rpmnet} or by optimization using stochastic gradient descent \cite{choy2020dgr}. This enables the definition of end-to-end learnable registration pipelines. Notably, the method by Yew and Lee \cite{yew2020rpmnet} additionally uses surface normals to compute Point Pair Features (PPF) as input. While there is effort to extract more robust features \cite{yew2020rpmnet,choy2020dgr}, these methods typically use secondary networks that predict the sharpness of the match matrix to deal with imperfect correspondences and outliers \cite{wang2019prnet, yew2020rpmnet}.

In contrast, another class of NN-based methods uses global features that represent whole point clouds and as such are more robust to imperfect correspondences. Seminal work in this direction by Aoki et al. \cite{aoki2019pointnetlk} poses iterative registration as the alignment of global features using an interpretation of the Lucas-Kanade algorithm. A deterministic formulation that replaces the approximate with an exact Jacobian is proposed in \cite{li2020deterministic}, which increases the stability of the approach. The method in \cite{sarode2019pcrnet} allows one-shot registration of global features. We show that global feature representations may be used as state representation in RL and learned jointly with the agent's policy.

\textbf{Reinforcement Learning in Object Pose Estimation:}
In the related domain of object pose estimation, Krull et al. \cite{krull2017poseagent} use RL to find a policy that efficiently allocates refinement iterations to a pool of pose hypotheses. Closely related to our approach, in RGB-based object pose estimation \cite{shao2020pfrl,busam2020moveit}, RL is used to train policies that manipulate an object's pose. Based on 2D segmentation masks, these agents learn to predict discrete refinement actions. In contrast, we focus on learning registration actions from 3D point clouds -- while RGB-based methods use pretrained optical flow estimation as state, we use siamese PointNets. Additionally, we integrate IL to quick-start training and stabilize RL.


\begin{figure*}
    \centering
    \includegraphics[width=0.7\textwidth]{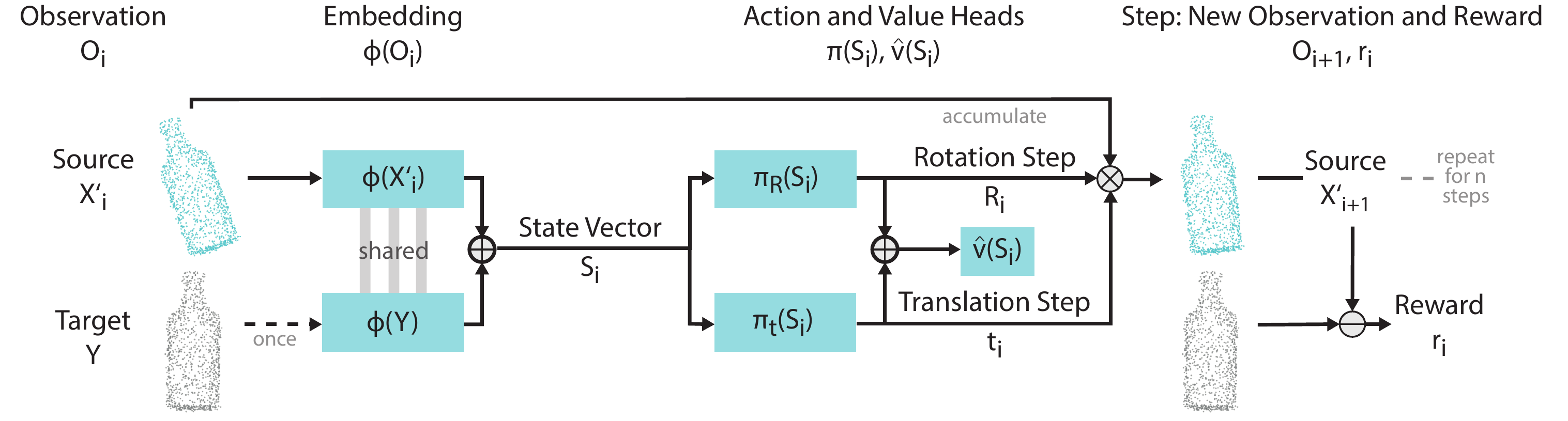}
    \caption{Architecture overview for one iteration of ReAgent.}
    \label{fig:architecture}
\end{figure*}

\section{Point Cloud Registration Agent}\label{sec:method}
In the following, we present our novel point cloud registration approach, based on imitation and reinforcement learning, called \textit{ReAgent}. For readers unfamiliar with point cloud registration, Section \ref{sec:sub_background} gives a brief introduction. In Section \ref{sec:sub_architecture}, we propose the fast and interpretable network architecture of the agent. Sections \ref{sec:sub_bc} and \ref{sec:sub_rl} present the learning procedure that enables accurate and robust registration. In Section \ref{sec:sub_implementation}, we discuss design choices that facilitate generalization to novel test categories on synthetic data and application to object pose refinement on real data.


\subsection{Background: Point Cloud Registration}\label{sec:sub_background}
Assume two point clouds, the \textit{source} $X$ and \textit{target} $Y$, that represent an object or scene surface are given. In the simplest case, both sets are identical but may in general only partially overlap. The observed source $X'$ is offset by an unknown rigid transformation $T' = [R'\in SO(3), t'\in \mathbb{R}^3]$, where $R'$ is a rotation matrix and $t'$ a translation vector. We define the tuple $O=(X', Y)$ as the \textit{observation}, where 
\begin{equation}
    X' = T' \otimes X.
\end{equation}
Given $O$, the task of point cloud registration is to find a rigid transformation $\hat{T}$ such that \begin{equation}
    \hat{T} \otimes X' = X.
\end{equation}
The optimal transformation would thus be $\hat{T}=T'^{-1}$ such that we retrieve the original alignment $(X,Y)$. In general, however, $\hat{T}$ is an error afflicted estimate of the registration. The target $Y$ guides this registration process.

When $n$ steps are taken to compute this transformation, this is referred to as \textit{iterative registration}. In every step $i$, a rigid transformation $\hat{T}_i$ is estimated and the observed source is updated by
\begin{equation}
    X'_{i} = \hat{T}_i \otimes X'_{i-1}.
\end{equation}
The goal is that the final estimate after $n$ steps is again
\begin{equation}\label{eq:iterative}
    X'_{n} = \hat{T}_{n} \otimes ... \otimes \hat{T}_1 \otimes X'_0 = \hat{T} \otimes X' = X.
\end{equation}

In a more general setting, one or both point clouds are noise afflicted. For example, their 3D coordinates may be jittered, their order may not correspond, not every point in the source may have a correspondence in the target and the number of points may not be identical. Such noise is typically due to the sensor recording the point clouds, varying view points or (self)occlusion.


\subsection{The ReAgent Architecture}\label{sec:sub_architecture}
The proposed architecture for our point cloud registration agent is shown in Figure \ref{fig:architecture}. The registration starts with a feature embedding that transfers the raw observed point clouds into global feature vectors. The concatenation of the source's and target's global feature vector is used as state representation, encoding the agent's information about the current registration state. A policy network uses the state representation to predict two action vectors, one for rotation and one for translation. Finally, the resulting transformation is applied to the observed source, iteratively improving the registration. In each such step, the agent receives a reward that judges how well it performs its task. The individual parts are now discussed in more detail.

\textbf{Learned state representation:}  The observed source and target may have varying shape and may be noise afflicted. Our goal is to learn a more robust and powerful representation than the bare point clouds. This is achieved by the feature embedding $\Phi(O)$, mapping from $N\times3$ dimensional observation to $1\times M$ dimensional state space. The source and target are passed through the embedding separately with shared weights. The concatenation of both global feature vectors is used as state $S$.

\begin{figure}
    \centering
    \includegraphics[width=0.9\linewidth]{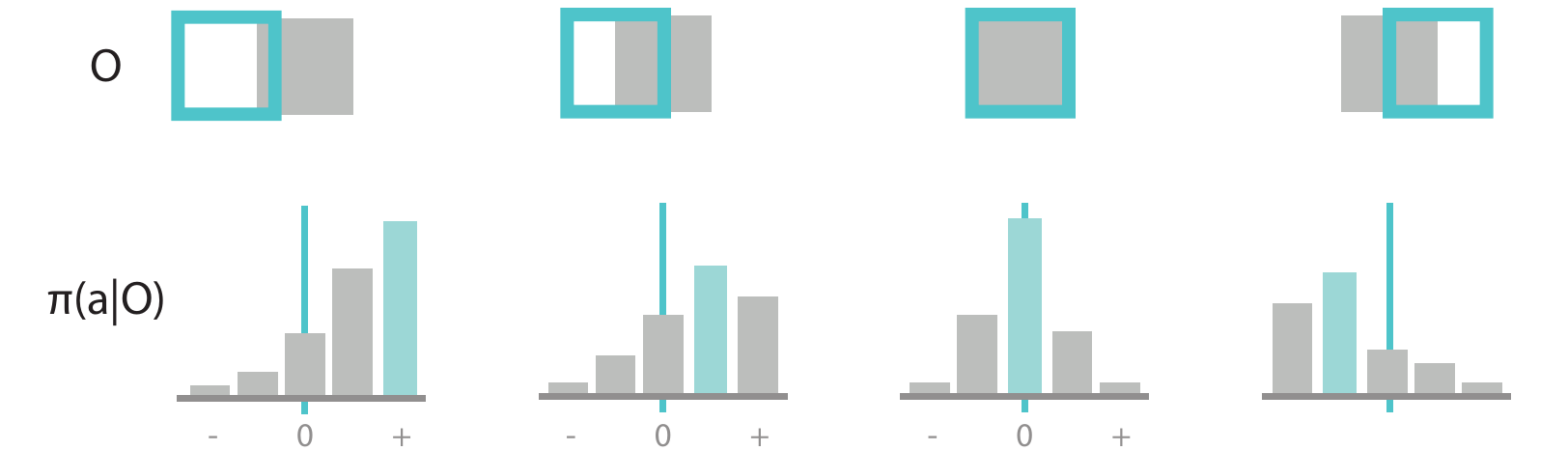}
    \caption{Illustration of interpretable actions. Top: Observed sources (cyan) with varying offset to the target (gray). Bottom: The probability of selecting each step size.}
    \label{fig:classification}
\end{figure}

\textbf{Discrete action space:} We observe that, when trying to reach an exact registration in every iteration (i.e, by repeated one-shot registration), a bad estimate in one step may lead to divergence of the whole registration process. To this end, related work proposes to robustify the matching process~\cite{wang2019prnet,yew2020rpmnet}. In an orthogonal approach, we aim to robustify the update steps themselves by using discrete, limited step sizes in each iteration. The discrete steps may be interpreted as the result of a classification of the observation into misalignment bins, as shown in Figure \ref{fig:classification}. Inspired by recent work by Shao et al. \cite{shao2020pfrl}, we use a set of discrete steps along and about each axis as action space. We propose to use an exponential scale for the step sizes to quickly cover a large space during initial registration, while allowing accurate fine-registration in later steps.

Given state $S$, the agent's policy $\pi(S)$ gives the probability of selecting action $a$. The policy is computed by the agent's action head and predicts the step sizes for the iteration. Note that $a$ is a vector of 6 sub-actions, one per rotation axis and translation axis. In addition, a value head estimates the baseline $\hat{v}(S)$. During the RL update, the baseline is subtracted from the returns of the actions to compute the advantage. This is commonly used to reduce variance as compared to using the returns directly.

\begin{figure}
    \centering
    \includegraphics[width=0.9\linewidth]{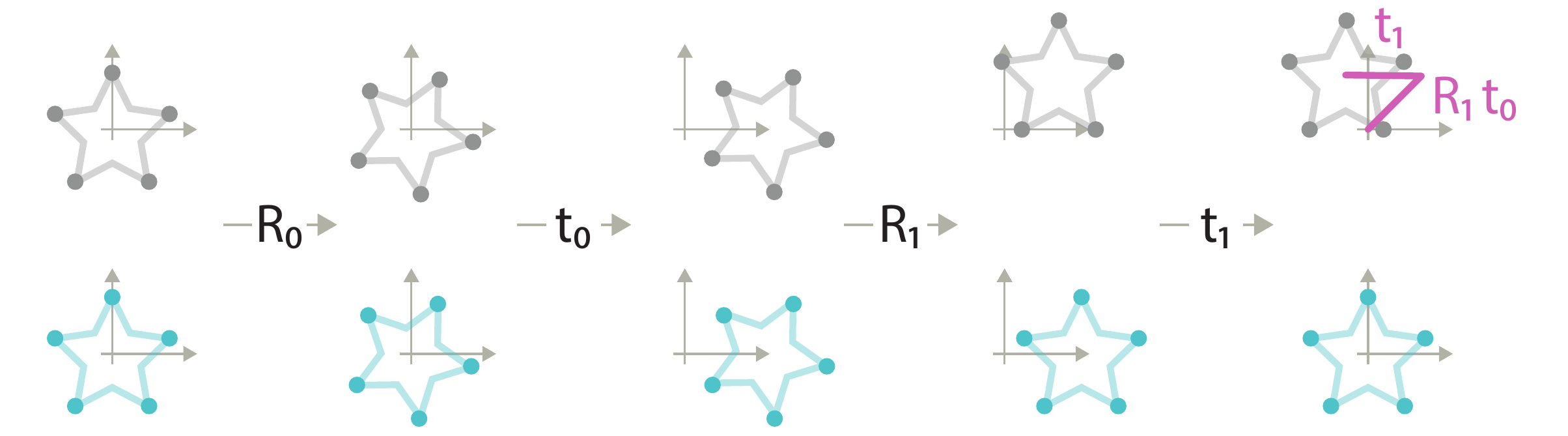}
    \caption{A transformation sequence and its effect on a point cloud using global (top) and disentangled transformation (bottom).}
    \label{fig:disentangled}
\end{figure}

\textbf{Disentangled transformation:} The concatenation of multiple rigid transformations with the source in Equation~\eqref{eq:iterative} may follow different conventions. The basic approach is to compute the matrix product of all $\hat{T}_i$ in homogenized form and apply this to $X'$. Yet, as shown in Figure \ref{fig:disentangled}, when the rotation center is not the origin, a rotation induces an additional translation of the point cloud since
\begin{equation}
    \begin{bmatrix}
        R_1 & t_1\\
        0   & 1
    \end{bmatrix}
    \begin{bmatrix}
        R_0 & t_0\\
        0   & 1
    \end{bmatrix} =
    \begin{bmatrix}
        R_1 R_0 & R_1 t_0 + t_1\\
        0   & 1
    \end{bmatrix}.
\end{equation}
Note that this is equal to iterative application of $\hat{T}_i$ to $X'$ as
\begin{equation}
    R_1 (R_0 X + t_0) + t_1 = R_1 R_0 X + R_1 t_0 + t_1.
\end{equation}
To support interpretability, however, we would like a rotation action to only result in a local rotation of the object. Moreover, we want the rotation and translation axes to align with the global coordinate axes such that an action in a specific axis always results in the same global displacement. Formally, we want iterative transformations to result in
\begin{equation}
    X_i = (\prod^i R_i) X + \sum^i t_i.
\end{equation}
Such disentanglement of rotation and translation not only benefits interpretability but, as shown for image-based object pose estimation by Li et al. \cite{li2018deepim}, is also beneficial for training of the agent; it does not need to account for the rotation-induced translation.

Following this idea, we propose a disentangled application of $\hat{T}$ to 3D point clouds. For iterative registration, we define the update rule for an accumulator $T_i=[R_i, t_i]$ by
\begin{equation}
    R_i = \hat{R}_i R_{i-1}, \quad t_{i} = \hat{t}_i + t_{i-1},
\end{equation}
which is initialized with $T_0=[I_{3\times3},0]$ and applied to the observed source by
\begin{equation}
    X'_i = R_i (X' - \mu_{X'}) + \mu_{X'} + t_i.
\end{equation}
Thereby, rotations are applied with the centroid of the observed source $\mu_{X'}$ as the origin. Since we only apply rigid transformations, the relation between points and the centroid does not change. 
The outcome is that no additional translation is introduced. This also holds when applying the accumulated transformation.


\subsection{Imitating an Expert Policy}\label{sec:sub_bc}
Learning a complex task, such as point cloud registration, from scratch using RL may take long to converge or may get stuck with a suboptimal policy; even more so if the state representation is learned jointly with the policy. To circumvent this issue, we initialize the state representation and the policy using IL.

In IL, the goal is to imitate the behavior of some domain expert. The simplest form of IL is Behavioral Cloning (BC). This assumes that, in every step, the agent has access to feedback from the expert. The feedback is used similarly to training data labels in supervised learning.


\textbf{Expert policy:} The expert feedback may come from interactions of a human expert or another algorithmic approach to solve the task. Since we can create training samples from point clouds by generating the initial rigid transformation, we have access to $T'$. We exploit this by defining two expert policies that reduce the true transformation error in the current step, given by
\begin{equation}
    \delta^R_i = R' R_i^\top, \quad \delta^t_i = t'_d - t_i,
\end{equation}
where $t'_d$ is the disentangled form of $t'$ that accounts for the translation induced by $R'$ using
\begin{equation}
    t'_d = t' - \mu_{X'} + R'\mu_{X'}.
\end{equation}
The expert policy either takes the largest possible step that reduces the \textit{absolute} error (greedy) or the \textit{signed} error (steady). The steady expert produces trajectories with monotonously decreasing error, while the greedy expert produces optimal trajectories at the cost of oscillation.

\textbf{Data gathering:} The initial transformation alone is, however, insufficient to train our agent. The agent will observe certain trajectories during inference that are not covered by the generated initial errors $T'$. To this end, we rollout trajectories by following the stochastic policy of the agent to gather a replay buffer. The distribution of training data in this buffer is more representative of what the agent will observe during inference and dynamically adapts as the agent improves. By using the stochastic policy, we also guarantee exploration. As the training converges, the entropy of the policy $H(\pi)$ -- and consequentially the exploration -- reduces. 

\textbf{Behavioral cloning:} The gathered training data, together with the annotation from the expert policy, allows us to train the agent using a 6-dimensional cross-entropy loss. For every observation, we gather registration trajectories. Once a certain number of trajectories is reached, the agent is updated using mini-batches from the shuffled buffer.


\subsection{Improving through Reinforcement}\label{sec:sub_rl}
The resulting agent policy is in two ways limited by the expert. On one hand, the agent cannot find a better policy than the expert as its actions would differ from the expert labels. On the other hand, different transformations will, in general, lead to different expert actions. If the observed sources are, however, indistinguishable due to symmetry, they might be represented by an identical state vector. This hinders training of the agent as this would require different actions to follow from the same state vector.

\textbf{Reward function:} The overall goal of the agent is to align source and target. Following the expert policy, this alignment will reflect the initial transformation $T'$. Rather, the alignment should be treated equally for equivalent transformations $\Tilde{T'}$ that result in indistinguishable observations where $X'\sim\Tilde{X'}$. Instead of training the agent to exactly imitate the expert policy, we thus additionally use RL and define the training objective by a reward function.

In the proposed RL task, equal consideration of equivalent transformations is achieved by using the mean Chamfer distance ($CD$) between the currently observed source $X'$ and true source $X=T'^{-1}X'$. This measure is insensitive to transformations that result in the same distance between closest points, e.g., rotations about a symmetry axis. Note, though, that the sampling rate of the point cloud may introduce fluctuations as $X'$ moves through undersampled regions of $X$ and vice-versa. However, this effect is lessened by considering the mean over all distances. 
Based on this, we define a step-wise reward
\begin{equation}
    r = \begin{cases}
    -\varepsilon^-,   &\quad CD(X'_i, X) > CD(X'_{i-1}, X)\\
    -\varepsilon^0,   &\quad CD(X'_i, X) = CD(X'_{i-1}, X)\\
    \varepsilon^+,    &\quad CD(X'_i, X) < CD(X'_{i-1}, X).\\
    \end{cases}
\end{equation}
Steps that reduce $CD$ are rewarded by $\varepsilon^+$, ``stop'' gets a negative penalty $-\varepsilon^0$ to discourage pausing and divergent steps are penalized by $-\varepsilon^-$. We choose $\varepsilon^- > \varepsilon^+$ to discourage alternating diverging and converging steps.

\textbf{Policy optimization:} The policy learned using IL already performs accurate registration. Large changes to the policy due to RL might result in forgetting and thereby worsening of the agent's performance. Rather, we want the policy after an RL update to be close to the previous policy. This is achieved by trust-region approaches such as Proximal Policy Optimization (PPO) \cite{schulman2017ppo}. The main idea of the clipped version of PPO is to limit the ratio between the previous and the updated policy by a fixed threshold. In addition, as observed in related work combining BC and GAIL~\cite{jena2020la}, it is benefitial to jointly optimize BC and RL objectives as to further limit divergence of the policy. 
In our combined approach, both IL and RL use the same replay buffer. Since the RL term considers equivalent transformations, the agent is able to differentiate between bad steps (discouraged by IL and RL), equivalent steps (discouraged by IL, encouraged by RL) and the best steps (encouraged by IL and RL).

\begin{algorithm}[t]
\caption{Combined Imitation and Reinforcement Learning using a Replay Buffer}
\begin{algorithmic}[1]\label{alg:learn}
\footnotesize
\definecolor{comment}{rgb}{.09,.75,.81}
\renewcommand{\algorithmicrequire}{\textbf{Input:}}
\FORALL{observations $O$ in $\mathcal{O}$}
    \STATE \textcolor{comment}{\% \textit{Gather replay buffer}}
    \FOR{$N$ trajectories}
        \FOR{$n$ refinement steps}
            \STATE agent predicts policy $\pi(O)$ and value $\hat{v}$
            \STATE action $a$ is sampled from policy $\pi(O)$
            \STATE take action $a$, receive reward $r$ and next $O'$
            \STATE add sample to buffer $b$, step observation $O$ = $O'$
        \ENDFOR
    \ENDFOR
    \STATE \textcolor{comment}{\% \textit{Process replay buffer}}
    \STATE compute return $R$, shuffle buffer $b$
    \FORALL{samples in buffer $b$}
        \STATE agent predicts new policy $\pi'(O)$ and value $\hat{v}'$
        \STATE \textcolor{comment}{\% \textit{Imitate expert}}
        \STATE expert predicts action $a^*$
        \STATE compute cross-entropy loss $l_{IL}$ of $\pi'(O)$ and $a^*$
        \STATE \textcolor{comment}{\% \textit{Reinforce}}
        \STATE compute PPO loss $l_{RL}$ of $\pi'(O)$ and $\pi(O)$
        \STATE \textcolor{comment}{\% \textit{Update agent}}
        \STATE $l$ = $l_{IL} + l_{RL} \cdot \alpha$
        \STATE backpropagate combined loss $l$
    \ENDFOR
    \STATE clear buffer $b$
\ENDFOR
\end{algorithmic} 
\end{algorithm}

\subsection{Implementation Details}\label{sec:sub_implementation}
The final combination of IL and RL that is used to train the agent is presented in Algorithm \ref{alg:learn}, where \textit{agent} implements the architecture shown in Figure \ref{fig:architecture}.

\textbf{Agent:} We choose a PointNet-like architecture \cite{qi2017pointnet} as feature embedding. As indicated by the findings of Aoki et al. \cite{aoki2019pointnetlk}, the T-nets in the original PointNet architecture are unnecessary for registration and are therefore omitted. We further observe that a reduced number of embedding layers is sufficient to learn an expressive embedding. The feature embedding $\Phi$ therefore reduces to 1D convolution layers of size $[64,128,1024]$, followed by max pooling as symmetric function. The concatenation of these $1024$ dimensional global features gives a $2048$ dimensional state vector.

In each iteration, the policy gives a step size for all 6 degrees of freedom. This is implemented as a prediction of the logits of a multi-categorical distribution. There is a total of 11 step sizes per axis: $[0.0033, 0.01, 0.03, 0.09, 0.27]$ in positive and negative direction, as well as a ``stop'' step. For rotation, step sizes are interpreted in radians. 

Shao et al. \cite{shao2020pfrl} propose to use shared initial layers for the action and value heads in an actor-critic design. We adapt this approach to our architecture and implement each head as fully-connected layers of size $[512, 256, D]$, where D is 33 for rotation and translation estimation and 1 for the value estimate. The concatenation of the middle layer of both action heads serves as input to the value head.

\textbf{Expert:} While the greedy policy achieves a lower error, when used to train the agent, both experts result in the same agent accuracy. We thus favor the more interpretable trajectories learned from the steady policy.

\textbf{PPO:} We use the PPO formulation from \cite{schulman2017ppo} for actor-critic architectures with an entropy term that encourages exploration. The advantage $\hat{A}$ in the PPO loss uses the agent's value estimate and Generalized Advantage Estimation (GAE) \cite{schulman2015gae}. For the reward function, we experimentally determine $(\varepsilon^+,\varepsilon^0,\varepsilon^-)=(0.5, 0.1, 0.6)$ to successfully guide the agent. 

\textbf{Hyperparameters:} Further parameters are the number of registration steps $n=10$, the number of trajectories per update $N=4$, the discount factor $\gamma=0.99$ and the GAE factor $\lambda=0.95$. The RL loss term is scaled by $\alpha=2$.

\textbf{Regularization:} While related methods use weight decay, batch or layer normalization for regularization \cite{aoki2019pointnetlk,wang2019dcp}, we observe that affine data augmentation achieved better results with our architecture. Namely, we use 1) random scaling sampled from $\mathcal{N}(1,0.1)$, clipped to $[0.5,1.5]$, 2) shearing in uniformly random direction by a random angle sampled from $\mathcal{N}(0,5)$, clipped to $[-15,15]\deg$ and 3) mirroring about a plane with uniformly random normal.


\begin{table*}
\setlength\tabcolsep{1ex}
\footnotesize
    \centering
    \begin{tabular}{l||c|c|c|c|c|c||c|c|c|c|c|c||c}
            & \multicolumn{6}{c||}{\textbf{Held-out Models}} & \multicolumn{6}{c||}{\textbf{Held-out Categories}} & \\
            &  \multicolumn{2}{c|}{MAE ($\downarrow$)} & \multicolumn{2}{c|}{ISO ($\downarrow$)} & ADI ($\uparrow$) & $\Tilde{CD}$ ($\downarrow$) &
               \multicolumn{2}{c|}{MAE ($\downarrow$)} & \multicolumn{2}{c|}{ISO ($\downarrow$)} & ADI ($\uparrow$) & $\Tilde{CD}$  ($\downarrow$) & T ($\downarrow$) \\
            & R         & t         & R         & t         & AUC       & $\times1e^{-3}$ &
              R         & t         & R         & t         & AUC       & $\times1e^{-3}$ & [ms] \\\hline
    ICP     & 3.59      & 0.028     & 7.81      & 0.063     &  90.6         & 3.49 &
              3.41      & 0.024     &  7.00     & 0.051     &  90.5         & 3.08 & \textbf{9} \\
        FGR$^+$     &  2.52       & 0.016     & 4.37      & 0.034     & 92.1          & 1.59 &
        1.68      & 0.011     &  2.94     & 0.024     & 92.7          & 1.24
        & 68 \\\hline
DCP-v2  &  3.48         & 0.025 &  7.01         & 0.052          & 85.8          & 2.52     & 
           4.51      & 0.031     & 8.89      & 0.064     & 82.3          & 3.74  & 23 \\
PointNetLK  & 1.64      & 0.012     & 3.33      &  0.026        & 93.0          &  1.03    & 
              1.61      & 0.013      & 3.22      &  0.028        & 91.6          &  1.51    & 45 \\\hline
ours IL & \textbf{1.46}      & \textbf{0.011}     & \textbf{2.82}     & \textbf{0.023}     &  \textbf{94.5}         & \textbf{0.75} & 
          1.38      & 0.010     &  2.59     & \textbf{0.020}     & \textbf{93.5}          & \textbf{0.95} & \multirow{2}{*}{21}\\
ours IL+RL  & 1.47      & \textbf{0.011}    & 2.87     & \textbf{0.023}  & 94.5 & \textbf{0.75} &
              \textbf{1.34}      & \textbf{0.009}     & \textbf{2.48}     & \textbf{0.020}  & 93.3 & 0.99 & \\
    \end{tabular}
    \caption{Results on ModelNet40 with held-out point clouds from categories 1-20 (left) and on held-out categories 21-40 (right). Note that $\downarrow$ indicates that smaller values are better. Runtimes are for a single registration with 1024 points per cloud. 
    $^+$ indicates that FGR additionally uses normals, while the remaining methods only use 3D coordinates.}.
    \label{tab:m40}
\end{table*}

\begin{figure*}
    \centering
    \includegraphics[width=\linewidth]{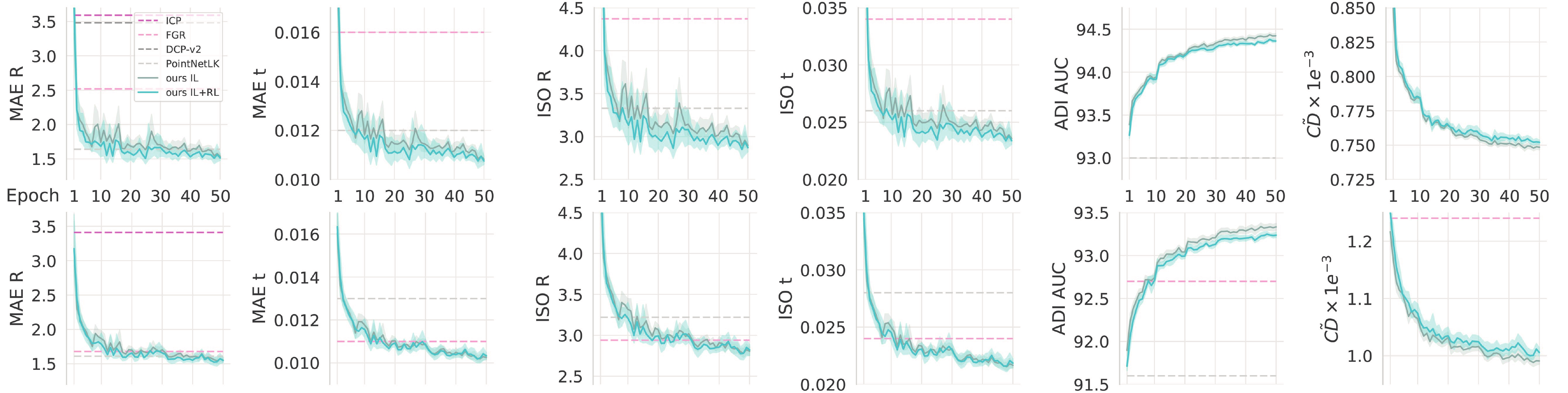}
    \caption{Convergence of ReAgent with 10 random seeds on held-out models (top) and categories (bottom) of ModelNet40. The lines show the mean and the shaded areas indicate the 95\%-confidence intervals. Best viewed digitally.}
    \label{fig:seeds}
\end{figure*}

\section{Experiments}\label{sec:results}
In the following, we evaluate the proposed point cloud registration agent on synthetic and real data. To evaluate our initial design goals of accuracy, inference speed and robustness, we consider noise-afflicted conditions on synthetic data. The generality of the approach is shown by results on held-out categories on synthetic data, the transfer to real data and by application to object pose refinement.

\subsection{Point Cloud Registration on Synthetic Data (ModelNet40)}\label{sec:poc}
As in prior work, we evaluate on ModelNet40~\cite{wu2015modelnet}, which features synthetic point clouds sampled from CAD models.

\textbf{Baselines:} For comparison, we evaluate two classical and two learning-based approaches. The former are Point-to-Point Iterative Closest Point (ICP)  \cite{besl1992icp} and Fast Global Registration (FGR) \cite{zhou2016fgr}, both as implemented in Open3D~\cite{open3d}. PointNetLK \cite{aoki2019pointnetlk} is an iterative approach based on global PointNet features. As with our approach, we set the number of iterations to 10. Deep Closest Point with Transformer (DCP-v2) \cite{wang2019dcp} is a local feature-based approach, predicting one-shot registration. As the latter methods provide no pretrained models on the ModelNet40 category splits, we retrain them using the published code. Note that all learning-based methods (including ours) use only 3D coordinates, while the FPFH features used by FGR additionally require surface normals. On ModelNet40, the models' normals are used.

\textbf{Metrics:} In line with prior work \cite{wang2019dcp,wang2019prnet}, we provide the Mean Average Error (MAE) over Euler angles and translations. Yew and Lee \cite{yew2020rpmnet} propose to additionally evaluate the isotropic error for rotation and translation (ISO), as well as a modified Chamfer distance ($\Tilde{CD}$) to cover symmetric ambiguity. The isotropic rotation error is computed by the geodesic distance between the rotation matrices and the isotropic translation error uses the Euclidean norm. All angles are given in degrees. Moreover, we provide the area under the precision-recall curve (AUC) for the Average Distance of Model Points with Indistinguishable Views (ADI) \cite{hinterstoisser2012adi}, a metric commonly used in object pose estimation. The ADI is normalized to the model diameter and we clip at a precision threshold of 10\% of the diameter. $\Tilde{CD}$ and ADI AUC implicitly consider symmetry.

\textbf{Training:} All methods are evaluated using an Intel Core i7-7700K and an NVIDIA GTX 1080. We train the proposed agent using Adam \cite{kingma2014adam} with AMSGrad \cite{reddi2019ams} and a batch size of 32. The replay buffer contains 4 trajectories of 10 steps each, resulting in a total of 1280 observations. We pretrain the agent for 50 epochs using IL ($\alpha=0$) on clean point clouds from ModelNet40. During pretraining, we start with a learning rate of $1e^{-3}$, and halve it every 10 epochs. We then fine-tune the policy for an additional 50 epochs on the first 20 categories of ModelNet40 with the noise defined in the following. Fine-tuning uses the same learning rate schedule, albeit starting from $1e^{-4}$. We provide separate results for training with only IL (\textit{ours IL}) and using the combined approach (\textit{ours IL+RL}). Note that this policy is used for all experiments and thus shows the generalization performance of the proposed approach.

%

\textbf{Results:}
To validate generalization to unseen points clouds and novel categories, we follow related work \cite{aoki2019pointnetlk,wang2019dcp,wang2019prnet} and use the point clouds generated by \cite{qi2017pointnet} based on ModelNet40. All approaches are trained on the training split of the first 20 categories. The data augmentations follow related work \cite{yew2020rpmnet}: Of the 2048 points, 1024 are randomly and independently subsampled for source and target to introduce imperfect correspondences. The source is transformed by a random rotation $R'$ of $[0,45]\deg$ per-axis and a random translation $t'$ of $[-0.5,0.5]$ per-axis. Random noise is sampled (again independently for source and target) from $\mathcal{N}(0, 0.01)$, clipped to $0.05$ and applied to the point clouds. Finally, the point clouds are shuffled as to permute the order of points. Table \ref{tab:m40} (left) shows results on the test split of the first 20 categories. Table \ref{tab:m40} (right) shows results on the test split of the second 20 categories. For consistency, we train the other learning-based approaches in the noisy condition.

As shown in Table \ref{tab:m40}, our approach successfully generalizes to novel point clouds and novel categories. We report improved accuracy across all metrics as compared to related work. The comparison in the rightmost column shows that our approach is also the fastest of the evaluated learning-based point cloud registration methods. Inference speed is even comparable to the one-shot method DCP-v2. However, the performance of DCP-v2 deteriorates with imperfect correspondences, as is the case with noisy observations.

When generalizing to held-out models, as shown in Figure \ref{fig:seeds}, the addition of RL successfully improves accuracy on the rotation-based metrics. As indistinguishable observations are due to rotations in this scenario these benefit most from the policy optimization. Yet, this improvement diminishes with novel categories, also indicated by the results on ScanObjectNN. Surprisingly, the consideration of symmetry via RL even slightly decreases mean performance on ADI AUC and $\Tilde{CD}$ over 10 random seeds.

\begin{table}[]
\footnotesize
\setlength\tabcolsep{1ex}
\centering
\begin{tabular}{c||r|c|r|c|c|r||c}
    \multirow{2}{*}{iter.}&  \multicolumn{2}{c|}{MAE ($\downarrow$)} & \multicolumn{2}{c|}{ISO ($\downarrow$)} & ADI ($\uparrow$) & $\Tilde{CD}$ ($\downarrow$) & $T_{c}$ \\
        & R         & t         & R         & t         & AUC   & $\times 1e^{-3}$ & [ms]  \\\hline
\rowcolor[rgb]{0.9,0.9,0.9} init& 22.35     & 0.238     & 44.49     & 0.478     & \hspace{1ex}3.4   & 225.6335 &  0        \\
    1   & 12.74     & 0.101     & 25.34     & 0.204     & 39.0  & 45.5217  &  3        \\
    2   & 6.31      & 0.050     & 12.17     & 0.100     & 69.9  & 10.7966  &  5        \\
\rowcolor[rgb]{0.9,0.9,0.9} ICP & 3.59 & 0.028 & 7.81 & 0.063 & 90.6 & 3.4882 & 9 \\
    3   & 3.32      & 0.025     & 6.46      & 0.052     & 83.7  & 2.9290   &  7        \\
    4   & 2.02      & 0.015     & 3.98      & 0.032     & 89.7  & 1.3366   &  9        \\
\rowcolor[rgb]{0.9,0.9,0.9} PNLK & 1.64 & 0.012 & 3.33 & 0.026 & 93.0 & 1.0305 & 45 \\
    5   & 1.50      & 0.011     & 2.97      & 0.024     & 92.8  & 0.9018   & 11        \\
    6   & 1.33      & 0.010     & 2.66      & 0.022     & 94.0  & 0.7814   & 13        \\
    7   & 1.33      & 0.010     & 2.64      & 0.022     & 94.3  & 0.7592   & 15        \\
    8   & 1.36      & 0.010     & 2.69      & 0.022     & 94.4  & 0.7528   & 17        \\
    9   & 1.40      & 0.010     & 2.76      & 0.023     & 94.7  & 0.7498   & 19        \\\hline
    10  & 1.47      & 0.011     & 2.87      & 0.023     & 94.4  & 0.7499   & 21        \\
\end{tabular}
\caption{Results per iteration for ReAgent (IL+RL) on held-out ModelNet40 models. See Table 1 (left) in the main text.}
\label{tab:stepbystep}
\end{table}

\textbf{Step-by-Step Results:}
Table \ref{tab:stepbystep} shows the results of ReAgent (IL+RL) after each iteration. $T_c$ indicates cumulative runtime for inference using a single observation. The conditions are identical to those in Table \ref{tab:m40} (left), using held-out ModelNet40 models. The step-by-step results show that the exponential scale allows to pick large step sizes to achieve a rough alignment initially. Within about 3 steps, our approach achieves an accuracy similar to ICP in less time. Smaller step sizes are used subsequently and the performance of, for example, PointNetLK is reached after about 5 steps and 11ms of runtime. We observe that, for the last steps, ReAgent further refines the Chamfer distance by aligning closer to an indistinguishable pose -- the error with respect to the true pose (indicated by MAE and ISO) increases slightly, while $\Tilde{CD}$ reduces. If accuracy with respect to the true rotation and translation is preferred over the symmetry-aware metrics, the number of iterations and thereby the runtime could be further reduced. For consistency across experiments, however, we report the results after 10 iterations in Table \ref{tab:m40}. Qualitative step-by-step examples are given in Figure \ref{fig:qualitative}. As shown, ReAgent quickly achieves a rough alignment within about 3 steps, which is then further refined using smaller step sizes.

\begin{figure*}
\centering
\begin{subfigure}{0.32\linewidth}
    \centering
    \includegraphics[width=\linewidth]{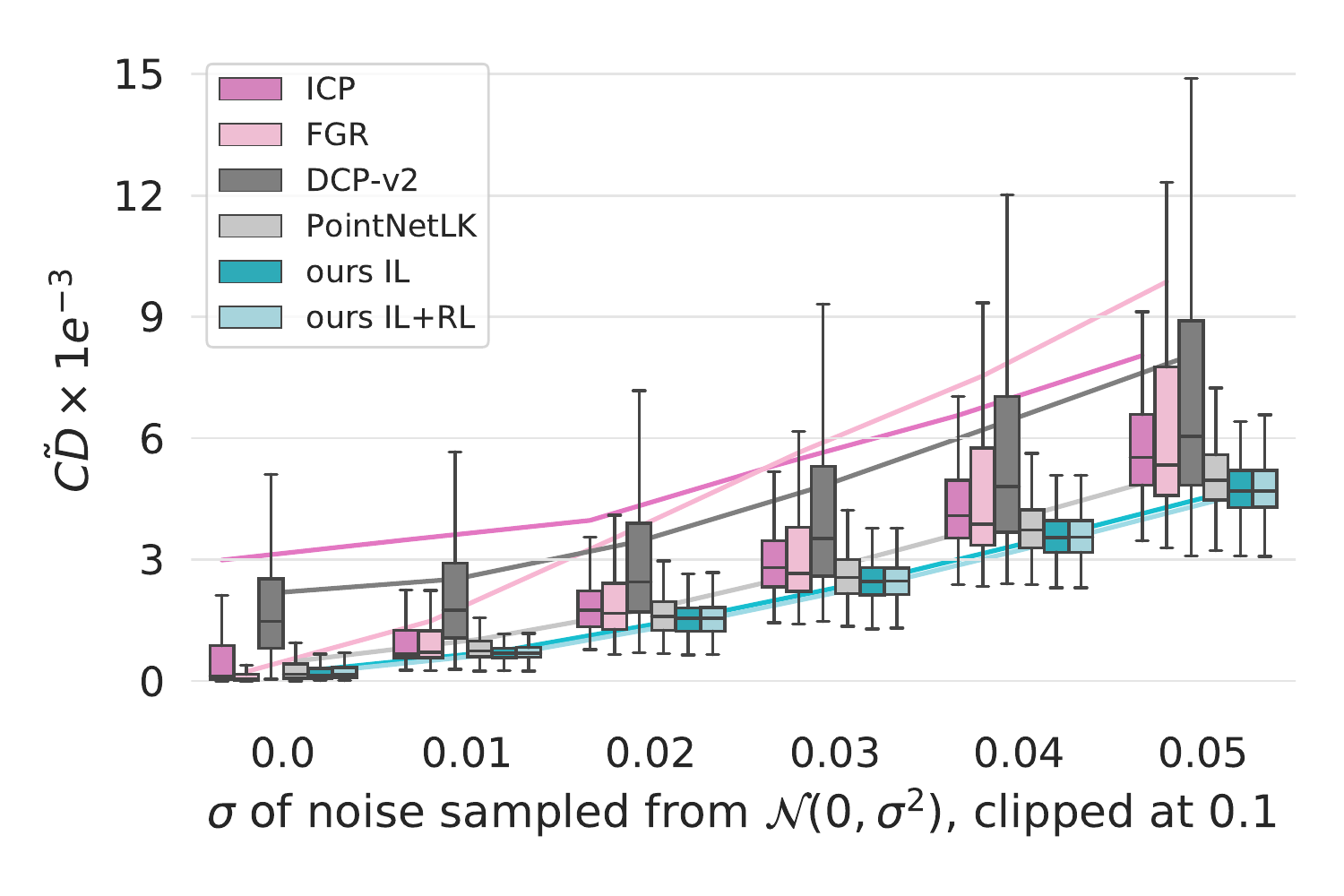}
    \caption{Varying noise magnitude.}
\end{subfigure}
\begin{subfigure}{0.32\linewidth}
    \centering
    \includegraphics[width=\linewidth]{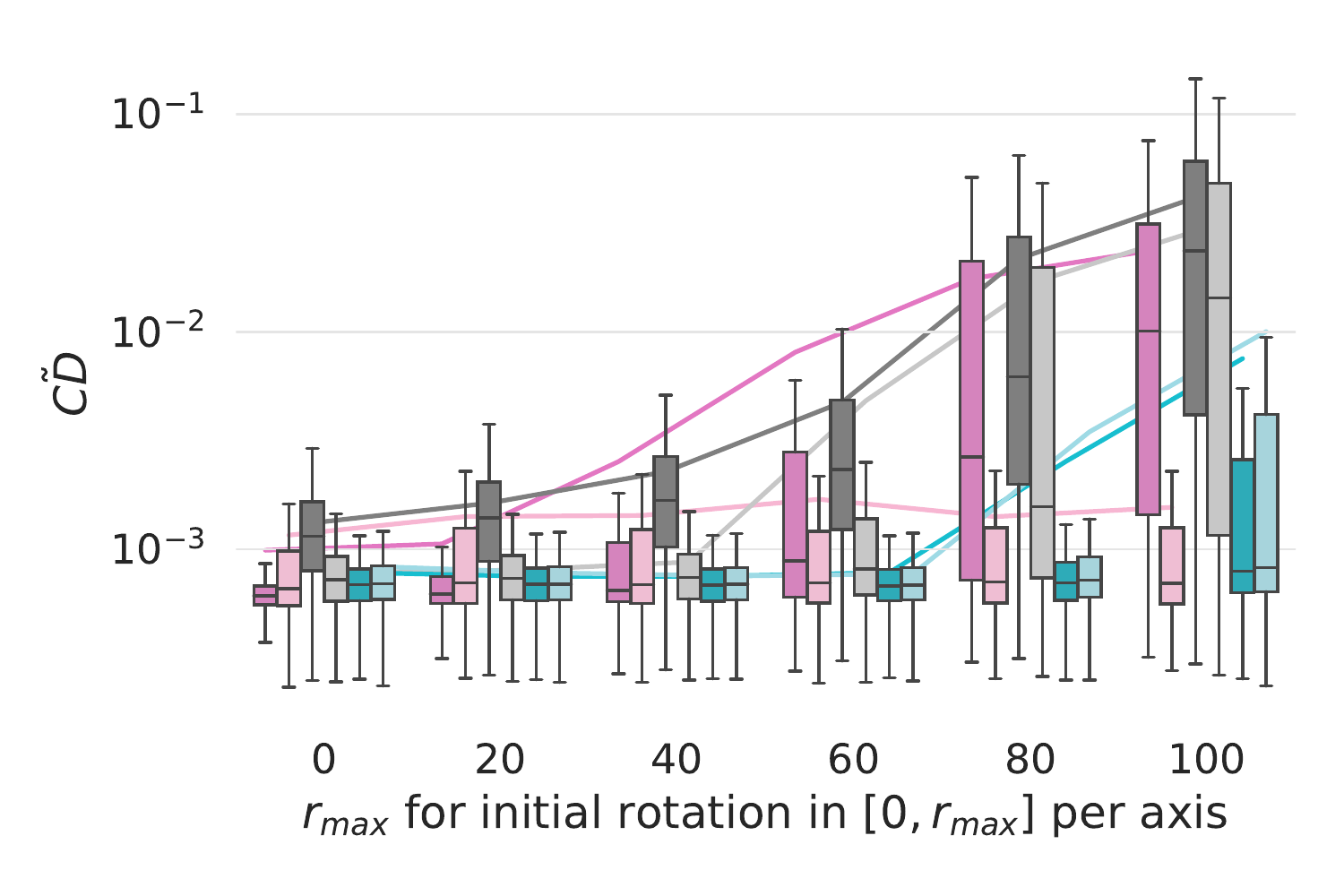}
    \caption{Varying initial rotation.}
\end{subfigure}
\begin{subfigure}{0.32\linewidth}
    \centering
    \includegraphics[width=\linewidth]{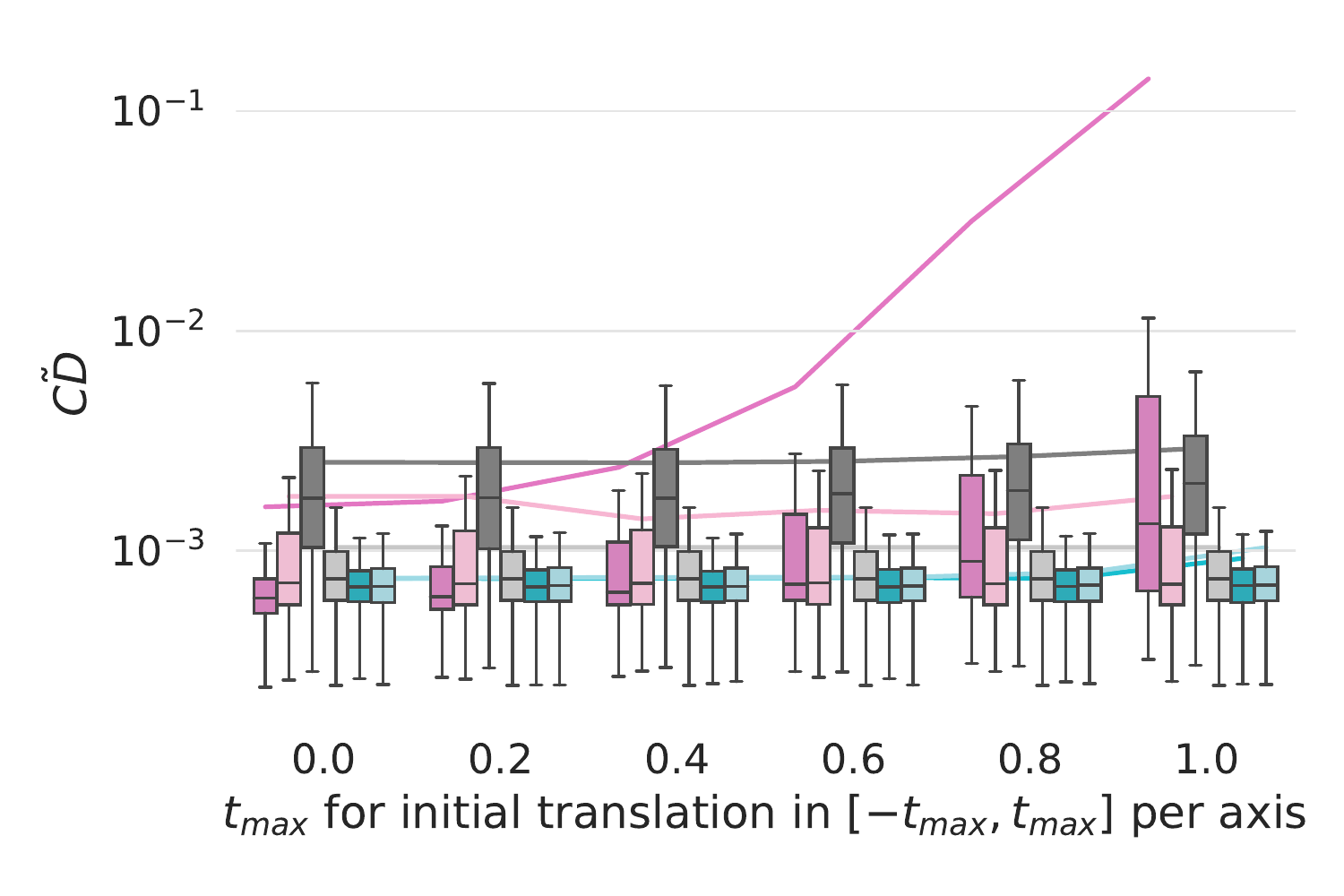}
    \caption{Varying initial translation.}
\end{subfigure}
\caption{Results on held-out ModelNet40 models with varying noise. The boxes show $[q_{.25},q_{.75}]$ and the median value. The whiskers indicate $[q_{.25} - 1.5 IQR,q_{.75} + 1.5 IQR]$, with $IQR = q_{.75}-q_{.25}$. The colored lines show the trend of the respective mean values. Note that large differences between mean and median are due to outliers.}
\label{fig:ablation}
\end{figure*}

\subsubsection{Ablation: Transformation and Noise Magnitude}
To show how the compared methods are affected by varying noise and initial conditions, we provide an ablation study in Figure \ref{fig:ablation}. We evaluate using the held-out ModelNet40 models as in Table \ref{tab:m40} (left).

\textbf{Noise Magnitude:} In Figure \ref{fig:ablation}a, the standard deviation of the distribution from which the noise is sampled is varied. The mean of the distribution remains constant at $0$ and clipping remains constant at $0.1$; increased from $0.05$ as compared to the main experiments.
FGR performs best in the noise-free condition but is heavily affected by increasing noise magnitude. Comparing the trend of the mean values, the other methods are similarly affected with ReAgent retaining overall best performance over all magnitudes.

\textbf{Initial Transformation:} In Figures \ref{fig:ablation}b-c, the respective component of the transformation is varied, while the other is kept as in the main experiments. The magnitude of the transformation is varied by increasing the upper-bound parameter of the transformation distributions.
Figure \ref{fig:ablation}b indicates that the PointNet-based methods are able to retain high accuracy within the range of initial rotations seen during training (up to $45\deg$ per axis). FGR, using the true surface normals of the models, is barely affected by increasing rotation magnitude.
In Figure \ref{fig:ablation}c, the performance of ICP is heavily affected by the translation magnitude. This is due to a fixed upper-bound for the correspondence distance of 0.5. ReAgent also shows a slight decrease in accuracy with the highest translation magnitude, as its step sizes are limited and thus more iterations need to be spent for initial alignment.


\begin{table}[]
\footnotesize
\setlength\tabcolsep{0.75ex}
    \centering
    \begin{tabular}{cc|cc|cc|cc||c}
    \multicolumn{2}{c|}{Data} & \multicolumn{2}{c|}{Expert} & \multicolumn{2}{c|}{$\Phi$, $\pi$} & \multicolumn{2}{c||}{$T$} & $\Tilde{CD}$\\
    \rotatebox{45}{stoch.} & \rotatebox{45}{augm.} & \rotatebox{45}{greedy} & \rotatebox{45}{steady} & \rotatebox{45}{deep} & \rotatebox{45}{wide} & \rotatebox{45}{basic} & \rotatebox{45}{disent.} & $\times1e^{-4}$ ($\downarrow$) \\\hline
               & \checkmark &            & \checkmark &            & \checkmark &            & global & 42.08 \\ 
    \checkmark &            &            & \checkmark &            & \checkmark &            & global & 11.81 \\ 
    \checkmark & \checkmark & \checkmark &            &            & \checkmark &            & global & 2.61 \\ 
    \checkmark & \checkmark &            & \checkmark & \checkmark &            &            & global & 2.92 \\
    \checkmark & \checkmark &            & \checkmark &            & \checkmark & \checkmark &        & 3.90 \\
    \checkmark & \checkmark &            & \checkmark &            & \checkmark &            & local  & 3.74 \\
    \checkmark & \checkmark &            & \checkmark &            & \checkmark &            & global & 2.63 \\
    \end{tabular}
    \caption{Ablation study. Results of \textit{ours IL}, pretrained on clean point clouds from held-out ModelNet40 categories.}
    \label{tab:design}
\end{table}

\subsubsection{Design of the Agent}
Table \ref{tab:design} presents results for central design choices. As shown, the use of the stochastic agent policy to gather additional training data (stoch.) is essential to the success of the method. To a lesser extent, the regularization by augmenting data through affine transformations (augm.) prevents overfitting. In Section \ref{sec:sub_bc}, the steady policy was already suggested to be more interpretable than the greedy policy. Even though the greedy policy on its own is more accurate than the steady policy, the results in Table \ref{tab:design} show that the agent trained using a steady policy achieves equally high accuracy. We support our choice to reduce the depth of the embedding (deep) and, instead, increasing the width of the head networks (wide) by the slightly increased accuracy. Finally, Table \ref{tab:design} highlights the importance of the representation and application of transformations. There is only a slight improvement by using a disentangled representation with rotations applied locally ($R_i=R_{i-1} \hat{R_i}$) as compared to using homogenized transformation matrices (basic). Using globally applied disentangled rotations ($R_i=\hat{R_i} R_{i-1}$), as suggested in Section \ref{sec:sub_architecture}, improves both accuracy and interpretability of our agent. By using a disentangled representation, the agent does not need to account for the rotation-induced translation. With global rotations, additionally, the rotation axes remain aligned with the global coordinate axes throughout trajectories.

\begin{figure}
    \centering
    \includegraphics[width=\linewidth]{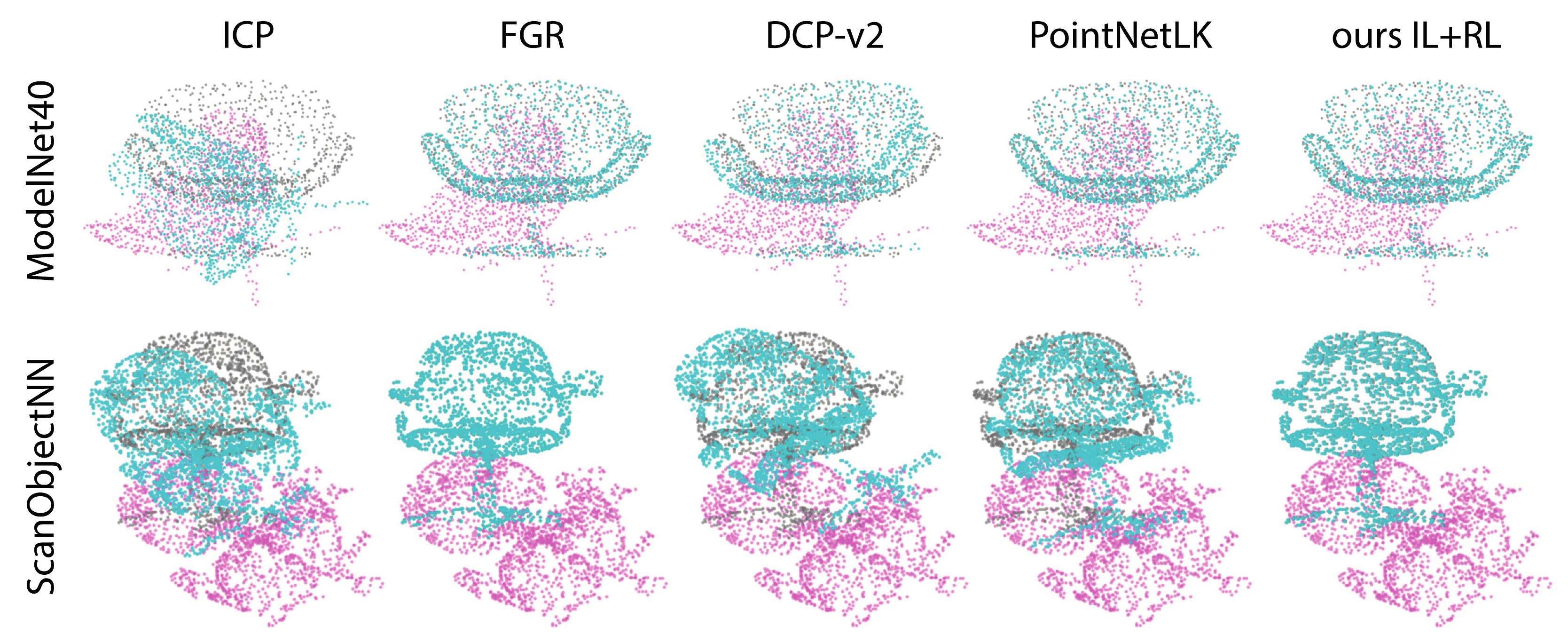}
    \caption{Qualitative examples. Columns show target (gray), initial (magenta) and registered source (cyan).}
    \label{fig:qualitative}
\end{figure}


\begin{table}
\setlength\tabcolsep{0.8ex}
\footnotesize
    \centering
    \begin{tabular}{l||c|c|c|c|c|c||c}
            & \multicolumn{6}{c||}{\textbf{Segmented Objects}}
            & \\
            &  \multicolumn{2}{c|}{MAE ($\downarrow$)} & \multicolumn{2}{c|}{ISO ($\downarrow$)} & ADI ($\uparrow$) & $\Tilde{CD}$ ($\downarrow$) 
               & T ($\downarrow$) \\
            & R         & t         & R         & t         & AUC       & $\times1e^{-3}$
              & [ms]\\\hline
    ICP     & 5.34      & 0.036     & 10.47     & 0.076      & 88.1      & 2.99
            & \textbf{19} \\
    FGR$^+$     & \textbf{0.11}      & \textbf{0.001}     & \textbf{0.19}      & \textbf{0.001}     & \textbf{99.7}      & \textbf{ 0.16}      
            & 131\\\hline
    DCP-v2  & 7.42      & 0.050     & 14.93          & 0.102          & 72.4          & 4.93      
            & 54 \\
PointNetLK  & 0.90      & 0.010     & 1.74          & 0.020          & 92.5          &  1.09        
            & 45\\\hline
ours IL     & 0.77      & 0.006     &  1.33     & 0.012     & 95.7      &  0.30 
            & \multirow{2}{*}{21} \\
ours IL+RL  & 0.93      & 0.007     &  1.66     & 0.014     & 95.4      &  0.34   
            & \\
    \end{tabular}
    \caption{Results on ScanObjectNN with the object segmented from the observation. Learning-based methods use the model trained on ModelNet40. 
    Note that $\downarrow$ indicates that smaller values are better. Runtimes are for a single registration and 2048 points per cloud. $^+$ indicates that FGR additionally uses normals.}
    \label{tab:son}
\end{table}


\subsection{Point Cloud Registration on Real Data (ScanObjectNN)}
To additionally evaluate generalization from synthetic to real data, we provide results on ScanObjectNN \cite{uy2019scanobjectnn}, featuring observations captured from an RGB-D sensor. We use the point clouds with segmented objects from ScanObjectNN dataset with 2048 points each. The same type of rigid transformations as in the previous condition are applied to the source. No additional noise is applied as the dataset already represents the characteristics of a specific depth sensor. For learning-based methods, the same models as in the previous conditions on ModelNet40 are used without any retraining or fine-tuning.

As shown in Table \ref{tab:son}, our approach transfers from training on ModelNet40 to testing on ScanObjectNN with high accuracy. Only FGR, additionally using normals to compute FPFH features, performs consistently better under this condition. However, inference time of FGR is almost 6 times higher compared to our approach. Notably, the inference time of PointNetLK and of our approach is barely affected by the doubling of the number of points. While DCP-v2 requires repeated neighborhood computation that negatively affects inference time, both PointNet-based approaches benefit from the independent embedding per point. Qualitative examples are shown in Figure \ref{fig:qualitative}.


\subsection{Application: Object Pose Refinement (LINEMOD)}
\begin{table*}
\newcolumntype{g}{>{\columncolor[rgb]{0.95,0.95,0.95}}c}
\setlength\tabcolsep{1ex}
\footnotesize
    \centering
    \begin{tabular}{l||gc|gcc|gccccc}
         & \multicolumn{2}{c}{RGBD-based} & \multicolumn{6}{c}{RGB-based} & \multicolumn{3}{c}{depth-based}\\
         
         & \multicolumn{2}{c}{$\overbrace{\hspace{2cm}}$} & \multicolumn{6}{c}{$\overbrace{\hspace{6.5cm}}$} & \multicolumn{3}{c}{$\overbrace{\hspace{2.5cm}}$}\\
         
         class & DenseFusion & DenseFusion ref.  & DPOD & DeepIM & DPOD ref.   & PoseCNN & PFRL & DeepIM & ICP-based   & ours IL & ours IL+RL\\\hline
         ape   & 79.5 & 92.3    & 53.3 & 78.7 & 87.7      & 27.8 & 60.5 & 77.0 & 79.1    & \textbf{97.2} & 96.9 \\
         vise  & 84.2 & 93.2    & 95.3 & 98.4 & 98.5      & 68.9 & 88.9 & 97.5 & 97.9    & \textbf{99.6} & \textbf{99.6} \\
         cam   & 76.5 & 94.4    & 90.4 & 97.8 & 96.1      & 47.5 & 64.6 & 93.5 & 93.5    & 99.0 & \textbf{99.3} \\
         can   & 86.6 & 93.1    & 94.1 & 97.6 & \textbf{99.7}      & 71.4 & 91.3 & 96.5 & 98.7    & 99.6 & 99.5 \\
         cat   & 88.8 & 96.5    & 60.4 & 85.2 & 94.7      & 56.7 & 82.9 & 82.1 & 96.0    & \textbf{99.8} & 99.7 \\
       driller & 77.7 & 87.0    & 97.7 & 91.6 & 98.8      & 65.4 & 92.0 & 95.0 & 84.2    & 98.8 & \textbf{99.0} \\
         duck  & 76.3 & 92.3    & 66.0 & 80.2 & 86.3      & 42.8 & 55.2 & 77.7 & 84.1    & \textbf{96.9} & 96.6 \\
\textit{eggbox}& \textbf{99.9} & 99.8    & 99.7 & 99.7 & \textbf{99.9}      & 98.3 & 99.4 & 97.1 & 98.4    & 99.8 & \textbf{99.9} \\
 \textit{glue} & 99.4 & \textbf{100.0}   & 93.8 & 99.5 & 96.8      & 95.6 & 93.3 & 99.4 & 99.1    & 99.2 & 99.4 \\
       puncher & 79.0 & 92.1    & 65.8 & 75.7 & 86.9      & 50.9 & 66.7 & 52.8 & 97.2    & 98.4 & \textbf{98.6} \\
         iron  & 92.1 & 97.0    & 99.8 & 99.7 & \textbf{100.0}     & 65.6 & 75.8 & 98.3 & 90.6    & 97.9 & 97.5 \\
         lamp  & 92.3 & 95.3    & 88.1 & 98.2 & 96.8      & 70.3 & 96.6 & 97.5 & 94.0    & \textbf{99.8} & 99.7 \\
         phone & 88.0 & 92.8    & 74.2 & 91.4 & 94.7      & 54.6 & 69.1 & 87.7 & 85.8    & 97.7 & \textbf{97.8} \\\hline
         mean  & 86.2 & 94.3    & 83.0 & 91.8 & 95.1      & 62.8 & 79.7 & 88.6 & 92.1    & \textbf{98.7} & \textbf{98.7}\\
    \end{tabular}
    \caption{Results on LINEMOD for $AD<0.1d$, initialized by DenseFusion (left), DPOD (mid) and PoseCNN (right).}
    \label{tab:lm}
\end{table*}

\begin{table}
\newcolumntype{g}{>{\columncolor[rgb]{0.95,0.95,0.95}}c}
\setlength\tabcolsep{1ex}
\footnotesize
    \centering
    \begin{tabular}{l|gcccc}
         class & PoseCNN & DeepIM & ICP-based   & ours IL & ours IL+RL \\\hline
         ape   &  5.2 & 48.6 & 38.0    & 70.6 & \textbf{71.7} \\
         vise  & 27.3 & 80.5 & 81.9    & 95.3 & \textbf{96.0} \\
         cam   & 12.5 & 74.0 & 56.1    & 87.7 & \textbf{89.6} \\
         can   & 26.2 & 84.3 & 81.2    & 95.7 & \textbf{95.8} \\
         cat   & 22.6 & 50.4 & 81.9    & 95.2 & \textbf{95.6} \\
       driller & 23.7 & 79.2 & 59.3    & 97.1 & \textbf{97.9} \\
         duck  &  9.9 & 48.3 & 50.0    & 65.0 & \textbf{69.4} \\
\textit{eggbox}& 73.9 & 77.8 & 93.1    & \textbf{99.1} & 98.9 \\
\textit{glue}  & 66.5 & 95.4 & 90.1    & \textbf{98.7} & 98.3 \\
       puncher & 13.0 & 27.3 & 64.7    & \textbf{91.3} & 90.1 \\
         iron  & 23.2 & 86.3 & 60.9    & \textbf{92.3} & 91.5 \\
         lamp  & 29.6 & 86.8 & 85.9    & 98.8 & \textbf{98.9} \\
         phone & 16.2 & 60.6 & 48.4    & \textbf{90.9} & \textbf{90.9} \\\hline
         mean  & 26.9 & 69.2 & 68.6    & 90.6 & \textbf{91.1} \\
    \end{tabular}
    \caption{Per-class results on LINEMOD for $AD < 0.05d$, initialized using PoseCNN.}
    \label{tab:lm_5perc}
\end{table}
\begin{table}
\newcolumntype{g}{>{\columncolor[rgb]{0.95,0.95,0.95}}c}
\setlength\tabcolsep{1ex}
\footnotesize
    \centering
    \begin{tabular}{l|gcccc}
         class & PoseCNN & DeepIM & ICP-based   & ours IL & ours IL+RL \\\hline
         ape   &  0.0 & 14.3 &  2.9    &  7.5 &  \textbf{9.0} \\
         vise  &  1.6 & 37.5 & 25.8    & 38.5 & \textbf{39.9} \\
         cam   &  0.5 & \textbf{30.9} &  4.2    & 17.8 & 24.8 \\
         can   &  1.0 & \textbf{41.4} & 10.6    & 39.7 & 41.3 \\
         cat   &  1.0 & 17.6 & 18.6    & \textbf{41.6} & 39.5 \\
       driller &  1.6 & 35.7 &  5.8    & 46.5 & \textbf{49.7} \\
         duck  &  0.3 & \textbf{10.5} &  3.5    &  6.8 &  6.9 \\
\textit{eggbox}& 17.9 & 34.7 & \textbf{73.3}    & 72.4 & 73.2 \\
\textit{glue}  & 15.4 & 57.3 & 41.9    & \textbf{76.4} & 74.1 \\
       puncher &  0.5 &  5.3 &  6.8    & \textbf{31.2} & 29.5 \\
         iron  &  0.7 & \textbf{47.9} &  5.0    & 34.2 & 34.9 \\
         lamp  &  1.6 & 45.3 & 44.0    & \textbf{67.8} & 66.9 \\
         phone &  0.8 & 22.7 &  4.5    & 24.6 & \textbf{25.7} \\\hline
         mean  &  3.3 & 30.9 & 19.0    & 38.8 & \textbf{39.6} \\
    \end{tabular}
    \caption{Per-class results on LINEMOD for $AD < 0.02d$, initialized using PoseCNN.}
    \label{tab:lm_2perc}
\end{table}

\begin{figure*}
\centering
\includegraphics[width=\linewidth]{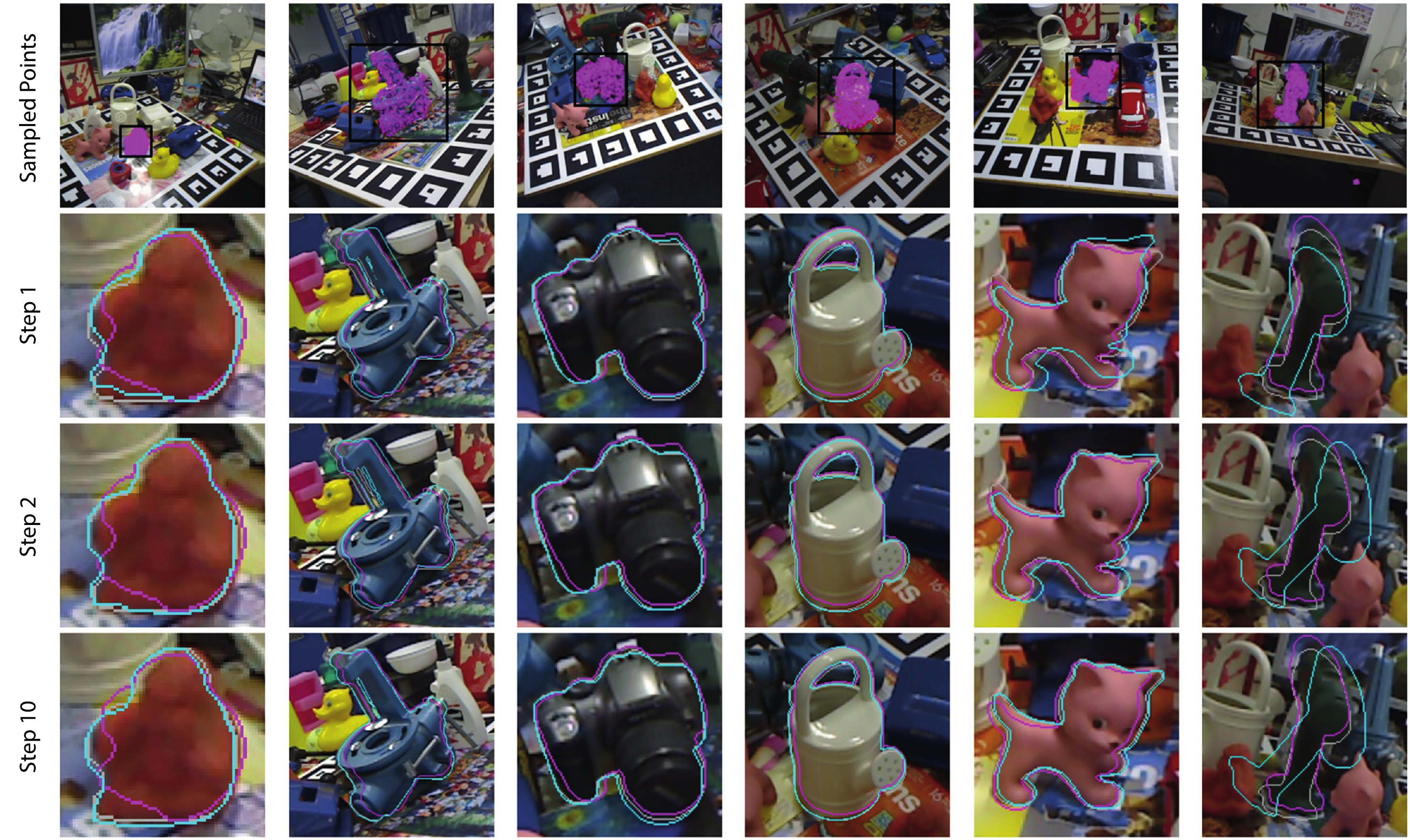}
\caption{Qualitative examples on LINEMOD using ReAgent (IL+RL). As shown in the top row, 1024 points are sampled within the estimated segmentation mask. The black box indicates the zoomed-in view. Outlines for target (gray), initial (magenta) and current source pose (cyan) are shown. The last column shows a failure case. Best viewed digitally.}
\label{fig:lm-qualitative-appendix}
\end{figure*}

In object pose estimation, the task equivalent to point cloud registration is referred to as \textit{object pose refinement}. Refinement is a step to significantly increase accuracy, starting from an initial pose provided by an object pose estimator. By evaluation on the LINEMOD dataset \cite{hinterstoisser2012adi}, commonly used in this domain, the performance of the presented approach in this real-world scenario is highlighted. LINEMOD consists of 15 objects of which 13 are used for evaluation.

\textbf{Baselines:} We compare our method to the reported performances of RGBD (DenseFusion \cite{wang2019dcp}), RGB-only (DPOD \cite{zakharov2019dpod}, DeepIM \cite{li2018deepim}, PFRL \cite{shao2020pfrl}) and depth-only (the ICP-based method used in \cite{xiang2017posecnn}) object pose refinement approaches. In each block in Table \ref{tab:lm}, the left-most column indicates the results of the method used for initialization (gray background). With our approach, we use the results provided for PoseCNN~\cite{xiang2017posecnn}. As in related work \cite{shao2020pfrl,li2018deepim}, we utilize the segmentation masks provided together with the initial poses.

\textbf{Metrics:} Hinterstoisser et al. \cite{hinterstoisser2012adi} propose two widely used evaluation metrics for object pose estimation. The Average Distance of Model Points (ADD) measures the distance between corresponding points under estimated and ground truth pose. To deal with symmetric objects (in italics in Table \ref{tab:lm}), the ADD with Indistinguishable Views (ADI) metric instead considers the distance between closest points. We indicate the mixed use of ADD and ADI by abbreviating with AD. The reported recall values are computed at precision thresholds of 10, 5 and 2\% of the object diameter.

\textbf{Training:} The training phase is slightly modified on LINEMOD. Instead of using a pre-training phase, we directly use the combined approach with a learning rate of $1e^{-3}$, halving it every 20 epochs for a total of 100 epochs. The influence of the PPO loss term is reduced to $\alpha=0.1$. Separate results for training using only IL are provided for comparison.

\textbf{Results:}  For training, we use the split defined in related work \cite{brachmann2016uncertainty,rad2017bb8,tekin2018real} and sample point clouds of size 1024 per training image as source. The sampling selects a random point on the object and finds its nearest neighbors in image space. $p\%$ of the points are sampled from the object (based on the ground-truth mask) and $100-p\%$ are sampled from the surrounding background, where $p$ is uniformly random in $[50,100\%]$. Thereby, we simulate partial observation of the object and imprecise segmentation (in place of affine augmentations). As target, we uniformly random sample 1024 points from the corresponding object model. The target point cloud is normalized to be mean centered and the farthest point to be of distance $1$. The same translation and scaling is applied to the source under ground-truth pose. As such, the distance from the origin provides an inductive bias on whether an (aligned) point belongs to the model or the background. Finally, we apply an uniformly random initialization error to the source, with the translation magnitude sampled from $[0,1]$ and the rotation magnitude sampled from $[0,90]$deg. During testing, we uniformly random sample 1024 points within the estimated segmentation mask and initialize the source using the estimated pose, as provided by the PoseCNN results \cite{xiang2017posecnn}.

As shown in Table \ref{tab:lm}, our approach outperforms all compared methods with respect to the mean AD and on most per-class results. Note that while it is more difficult for RGB methods to estimate an accurate z-translation from the camera, they more easily recover from bad segmentation masks -- and vice-versa for depth-based methods. For those methods that provide results using stricter AD thresholds, we additionally provide a comparison in Tables \ref{tab:lm_5perc} and \ref{tab:lm_2perc}. Again, our approach increases accuracy by several percent, even achieving accuracy on the stricter $0.05d$ threshold that is competitive to the performance of the compared methods on the permissive $0.1d$ threshold. The results, moreover, indicate that the addition of RL is especially beneficial for these stricter thresholds. Since we train a single model and do not provide the agent with any information on the object class, prioritizing features that support refinement of one class might hinder that of another one. Sacrificing some generality by introducing class labels as input could thus increase performance.

For the step-wise illustrations in Figure~\ref{fig:lm-qualitative-appendix}, we zoom-in on the objects, as indicated by the black box in the top row, to highlight the high pose accuracy achieved by our ReAgent, barely distinguishable from the ground-truth pose. The pose accuracy already increases significantly within the first few ReAgent steps. For consistency, we keep using 10 steps on LINEMOD (as in the ModelNet40 experiments). While this might be reduced in real-world applications to speed-up refinement even more (below the 22ms achieved at the moment with 10 steps), the higher number of steps enables increased robustness to initialization errors.

A runtime comparison is more difficult on LINEMOD as we rely on reported numbers using different hardware. Still, the 22ms required by our approach (GTX 1080) compares favorably to 30ms for DPOD refinement (Titan X), 83ms for DeepIM (GTX 1080 Ti) and PFRL with 240ms (RTX 2080 Ti). Wang et al. \cite{wang2019densefusion} report no hardware and only provide runtimes for scenes of 3 to 6 objects, with 20+10ms for DenseFusion refinement (20ms for initial embeddings) and 10.4s for the ICP-based method from \cite{xiang2017posecnn}. The latter refines multiple hypotheses and uses rendering-based verification.


\section{Conclusion}\label{sec:conclusion}
We present a novel point cloud registration agent, called ReAgent, that leverages imitation and reinforcement learning to achieve accurate and robust registration of 3D point clouds. Its discrete actions and steady registration trajectories are shown to be interpretable, while achieving fast inference times. The generality of our approach is evaluated in experiments on synthetic and real data. On the synthetic ModelNet40 dataset, our approach outperforms all evaluated classical and learning-based state-of-the-art registration methods. On ScanObjectNN, featuring real data, our approach achieves state-of-the-art for all comparable methods that only use 3D coordinates. While introducing normals, such as in FGR, achieves slightly better registration accuracy, our method is 6 times faster, making it more suitable for real-time applications. Finally, on LINEMOD, our approach achieves state-of-the-art accuracy on the object pose estimation task, outperforming competing pose refinement methods. We believe that the discussed properties make ReAgent suitable to be used in many application domains, such as object pose estimation or scan alignment.

To improve performance in object pose estimation, a semantic segmentation head as proposed in the original PointNet paper \cite{qi2017pointnet} may be adapted to iteratively improve object segmentation during refinement. In addition, the combination of ReAgent with rendering-based verification such as proposed in \cite{bauer2020verefine} should be explored to efficiently consider multiple initial pose estimates.
Further applications of the proposed method, such as self localization in large maps, will require efficient means to determine a state embedding from vast numbers of points. Replacing the PointNet embedding with (Deep) Lean Point Networks \cite{le2020lpn} would increase model capacity, allowing transfer to such complex domains. Furthermore, discrete steps induce finite accuracy, bound by the smallest step size. Similarly, the largest step size bounds the initial error that may be overcome by the agent in a given number of iterations. To further generalize our approach, a dynamically predicted scale factor could adapt step sizes.

\textbf{Acknowledgements:} This work was supported by the TU Wien Doctoral College TrustRobots and the Austrian Science Fund (FWF) under grant agreements No. I3968-N30 HEAP and No. I3969-N30 InDex. We would like to thank Matthias Hirschmanner, Kiru Park and Jean-Baptiste Weibel for valuable feedback and insightful discussions.


{\small
\bibliographystyle{plainnat}
\bibliography{bibliography}
}

\section{Appendix}

\subsection{Architecture Details}\label{sec:architecture}
Table \ref{tab:architecture} details the architecture with all used layers, their input and output dimensions. Note that the initial embedding is computed for both source and target with shared weights. Also, the action embeddings and policies are computed for both rotation and translation, although the layers are given only once in Table \ref{tab:architecture}.

\begin{table}[h]
\footnotesize
\centering
\begin{tabular}{c|cc}
     Layer  & In                & Out \\\hline
     \multicolumn{3}{c}{\cellcolor[rgb]{0.9,0.9,0.9}Embeddings $\phi(X'_i)$ and $\phi(Y)$ (shared)} \\\hline
     Conv1d & $N\times 3$       & $N\times 64$   \\
     ReLU   &                   &   \\
     Conv1d & $N\times 64$      & $N\times 128$  \\
     ReLU   &                   &   \\
     Conv1d & $N\times 128$     & $N\times 1024$ \\
     max    & $N\times 1024$    & $1\times 1024$ \\\hline
     \multicolumn{3}{c}{\cellcolor[rgb]{0.9,0.9,0.9}State $S$ via $\phi(X'_i) \oplus \phi(Y)$} \\\hline
     concat & $2 (1\times 1024)$& $2048$ \\\hline
     \multicolumn{3}{c}{\cellcolor[rgb]{0.9,0.9,0.9}Embeddings $\phi_R(S)$ and $\phi_t(S)$} \\\hline
     FC     & $2048$            & $512$ \\
     ReLU   &                   &   \\
     FC     & $512$             & $256$ \\
     ReLU   &                   &   \\\hline
     \multicolumn{3}{c}{\cellcolor[rgb]{0.9,0.9,0.9}Value $\hat{v}$, using $\phi_R(S) \oplus \phi_t(S)$} \\\hline
     concat & $2 (256)$         & $512$ \\
     FC     & $512$             & $256$ \\
     ReLU   &                   &   \\
     FC     & $256$             & $1$ \\\hline
     \multicolumn{3}{c}{\cellcolor[rgb]{0.9,0.9,0.9}Policies $\pi_R(\phi_R(S))$ and $\pi_t(\phi_t(S))$} \\\hline
     FC     & $256$             & $33$ \\
     reshape& $33$              & $3\times 11$\\
\end{tabular}
\caption{ReAgent network architecture.}
\label{tab:architecture}
\end{table}
\subsection{Definition of Used Metrics}\label{sec:metrics}
The metrics provided for our experiments are commonly used in related work \cite{wang2019dcp,yew2020rpmnet,hinterstoisser2012adi}. We provide their definition in condensed form in the following section.

\textbf{Mean Absolute Error (MAE):} 
The MAE between a vector of estimated $v'$ and true values $v$ is defined as
\begin{equation}
\footnotesize
    MAE_v = \frac{1}{3} \sum |v' - v|,
\end{equation}
where $v$ is either the vector of Euler angles in degrees representing the rotation or the translation vector.

\textbf{Isotropic Error (ISO):} While MAE considers axes individually, ISO is computed over the full rotation and full translation. For the rotation error $ISO_R$, we compute the angle of the residual rotation matrix by
\begin{equation}
\footnotesize
    ISO_R = \arccos \frac{trace(R'^{-1}, R) - 1}{2}
\end{equation}
and Euclidean distance between the estimated and true translation
\begin{equation}
\footnotesize
    ISO_t = ||t' - t||_2.
\end{equation}

\textbf{Chamfer Distance ($\Tilde{CD}$):} The Chamfer distance is used in the reward function of ReAgent. It is defined as
\begin{equation}
\footnotesize
    CD(X, Y) = \frac{1}{|X|} \sum_{x\in X} \min_{y\in Y} ||x-y||_2^2.
\end{equation}

\textbf{Modified Chamfer Distance $\Tilde{CD}$:} A modified variant is proposed in \cite{yew2020rpmnet}. Compared to MAE and ISO, it considers the distances between points and not the transformations. It therefore implicitly considers symmetry. Based on the definition of the Chamfer distance, it is defined as
\begin{equation}
\footnotesize
\Tilde{CD}(X, Y) = CD(X, Y_{clean}) + CD(Y, X_{clean}),
\end{equation}
where $X_{clean}$ and $Y_{clean}$ are the respective point cloud before applying noise and we compute $\Tilde{CD}(X', Y)$.

\textbf{Average Distance of Model Points (ADD) and Average Distance of Model Points with Indistinguishable Views (ADI):}
The ADD is proposed in \cite{hinterstoisser2012adi}. Given a model under an estimated pose $X'$ and under the true pose $Y$, it is defined as the mean distance between corresponding points
\begin{equation}
\footnotesize
    ADD = \frac{1}{|Y|} \sum_{y\in Y,x'\in X'} ||y - x'||_2.
\end{equation}
In addition, Hinterstoisser et al. \cite{hinterstoisser2012adi} propose to account for symmetrical true poses by considering the closest point pairs, defined as
\begin{equation}
\footnotesize
    ADI = \frac{1}{|Y|} \sum_{y\in Y} \min_{x'\in X'} ||y - x'||_2.
\end{equation}
The ADI recall for a specific precision threshold and $N$ test samples is
\begin{equation}
\footnotesize
    ADI_{th} = \frac{1}{N} \sum_i \begin{cases}
       0, & ADI_i > th\\
       1, & ADI_i \leq th,
    \end{cases}
\end{equation}
where $ADI_i$ is the ADI of the $i^{th}$ test sample. The ADD recall is computed analogously. Note that, as $\Tilde{CD}$, the ADI metric implicitly considers symmetry.

\textbf{ADI Area-under-Curve (ADI AUC):} $ADI_{th}$ is computed at uniformly spaced thresholds up to maximum precision threshold. This results in a monotonically increasing precision-recall curve. ADI AUC is then defined as the area under this curve
\begin{equation}
\footnotesize
    AUC = \frac{1}{th_{max}} \sum_{th\in[0:\Delta:th_{max}]} ADI_{th} \cdot \Delta,
\end{equation}
where $\Delta$ is the threshold spacing. We use $th_{max}=0.1d$ and $\Delta=1e^{-3}d$, where $d$ is the diameter of the point cloud computed as maximal distance between any two points
\begin{equation}
\footnotesize
    d = \max_{x_1,x_2 \in X} ||x_1 - x_2||_2.
\end{equation}

\end{document}